\def\year{2019}\relax
\documentclass[letterpaper]{article} 
\usepackage{aaai19}  
\usepackage{times}  
\usepackage{helvet}  
\usepackage{courier}  
\usepackage{url}  

\usepackage{graphicx}  
\newcommand{\citet}[1]
{\citeauthor{#1}~\shortcite{#1}}
\newcommand{\citep}{\cite}

\usepackage{xcolor,colortbl} 
\usepackage{amsmath} 
\usepackage{amsfonts}
\usepackage{amssymb}
\usepackage{array}
\usepackage{booktabs}
\usepackage{siunitx}
\usepackage{makecell}
\usepackage{multirow}

\DeclareMathOperator*{\argmax}{arg\,max}

\begin{document}
%

\title{Generating Distractors for Reading Comprehension Questions \\ from Real Examinations}
\author{Yifan Gao,\textsuperscript{1}\thanks{This work was mainly done when Yifan Gao was an intern at Tencent AI Lab.} Lidong Bing,\textsuperscript{2}\thanks{This work was mainly done when Lidong Bing was working at Tencent AI Lab.} Piji Li,\textsuperscript{3} Irwin King,\textsuperscript{1} Michael R. Lyu\textsuperscript{1}  \\
	{\textsuperscript{1} Department of Computer Science and Engineering, }\\
	{The Chinese University of Hong Kong, Shatin, N.T., Hong Kong}\\
	{\textsuperscript{2} R\&D Center Singapore, Machine Intelligence Technology, Alibaba DAMO Academy}\\
	{\textsuperscript{3} Tencent AI Lab}\\
    { \textsuperscript{1}\{yfgao,king,lyu\}@cse.cuhk.edu.hk}
    { \textsuperscript{2}l.bing@alibaba-inc.com} 
	{ \textsuperscript{3}pijili@tencent.com}
}

\maketitle
\begin{abstract}
We investigate the task of distractor generation for multiple choice reading comprehension questions from examinations. In contrast to all previous works, we do not aim at preparing words or short phrases distractors, instead, we endeavor to generate longer and semantic-rich distractors which are closer to distractors in real reading comprehension from examinations. Taking a reading comprehension article, a pair of question and its correct option as input, our goal is to generate several distractors which are somehow related to the answer, consistent with the semantic context of the question and have some trace in the article. We propose a hierarchical encoder-decoder framework with static and dynamic attention mechanisms to tackle this task. Specifically, the dynamic attention can combine sentence-level and word-level attention varying at each recurrent time step to generate a more readable sequence. The static attention is to modulate the dynamic attention not to focus on question irrelevant sentences or sentences which contribute to the correct option. Our proposed framework outperforms several strong baselines on the first prepared distractor generation dataset of real reading comprehension questions. For human evaluation, compared with those distractors generated by baselines, our generated distractors are more functional to confuse the annotators.
\end{abstract}

\section{Introduction}



Reading comprehension (RC) is regarded as an avant-garde task in NLP research for practising the capability of language understanding. 
Models with recent advances of deep learning techniques are even capable of exceeding human performance in some RC tasks, such as for questions with span-based answers \cite{Yu2018QANetCL}. 
However, it is not the case when directly applying the state-of-the-art models to multiple choice questions (MCQs) in RACE dataset  \cite{Lai2017RACELR}, elaborately designed by human experts for real examinations, where the task is to select the correct answer from a few given options after reading the article. The performance gap between the state-of-the-art deep models (53.3\%) \cite{Tay2018MultirangeRF} and ceiling (95\%) \cite{Lai2017RACELR} is significant. 
One possible reason is that in MCQs, besides the \textit{question} and the correct \textit{answer} option, there are a few \textit{distractors} (wrong options) to distract humans or machines from the correct answer.
Most distractors are somehow related to the answer and consistent with the semantic context of the question, and all of them have correct grammar \cite{goodrich1977distractor,Liang2018DistractorGF,Ma2010DiversifyingQS}. Furthermore, most of the distractors have some trace in the article, which fails the state-of-the-art models utilizing context matching only to yield decent results.
\begin{figure}
    \centering
    \includegraphics[width=1.0\columnwidth]{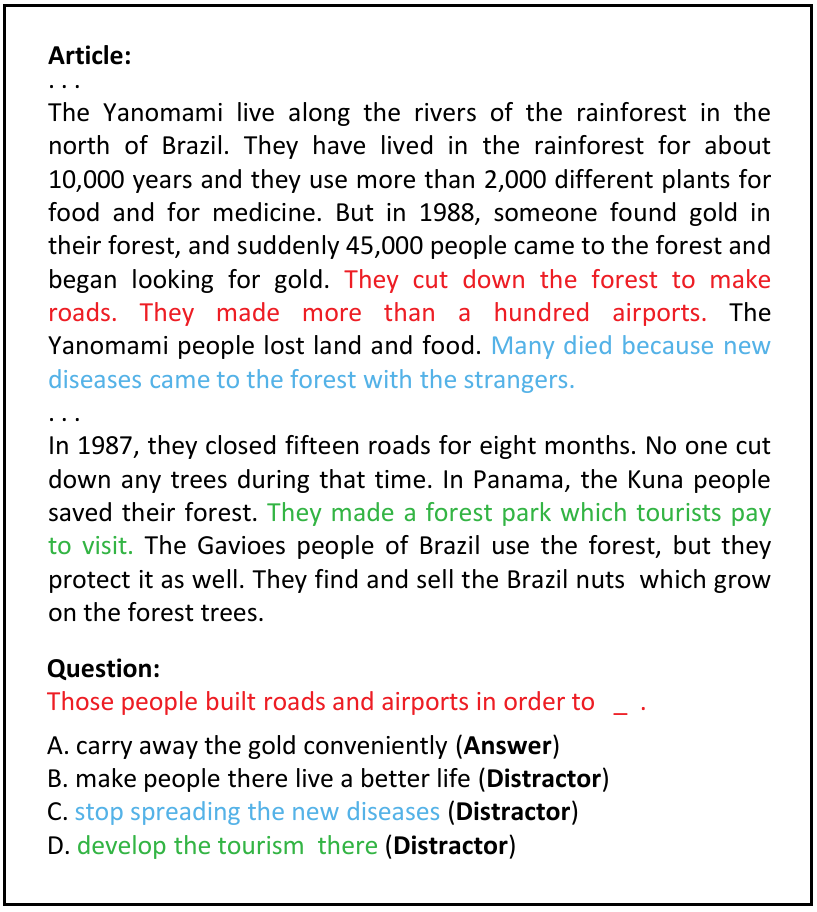}
    \caption{Sample multiple choice question along with the corresponding article. The question, options and their relevant sentences in the article are marked with the same color.}
    \label{fig:example}
\end{figure}

The MCQs in the RACE dataset are collected from the English exams for Chinese students from grade 7 to 12.
Constructing RACE-like MCQ dataset is important and nontrivial, because poor distractor options can make the questions almost trivial to solve \cite{Welbl2017CrowdsourcingMC} and reasonable distractors are time-consuming to design. 
In this paper, we investigate the task of automatic \textit{distractor generation} (DG) . The task aims to generate reasonable distractors for RACE-like MCQs, given a reading comprehension article, and a pair of question and its correct answer originated from the article.
Figure \ref{fig:example} shows an example multiple choice question with four options. We can find that all options are grammatically coherent with the question, and semantically relevant to the article. 
Distractor generation is of great significance in a few aspects. It can aid the preparation of MCQ reading comprehension datasets. With large datasets prepared, it is expectable that the performance of reading comprehension systems for MCQs will be boosted, as we have observed such improvements \cite{Yang2017SemiSupervisedQW} by applying generated question-answer pairs to train models to solve SQuAD questions. 
It could also be helpful to alleviate instructors' workload in designing MCQs for students.

Automatic DG is different from previous distractor preparation works, which basically follow an extraction-selection manner. 
First, a distractor candidate set is extracted from multiple sources, such as GloVe vocabulary \cite{Pennington2014GloveGV}, noun phrases from textbooks \cite{Welbl2017CrowdsourcingMC} and articles \cite{Araki2016GeneratingQA}.
Then similarity based \cite{Guo2016QuestimatorGK,Stasaski2017MultipleCQ,Kumar2015RevUPAG,Mitkov2003ComputeraidedGO,Zhou2012ACA,Yang2011EfficientSG} or learning based \cite{Liang2018DistractorGF,Sakaguchi2013DiscriminativeAT,Liang2017DistractorGW,Yang2010OnlineLF} algorithms are employed to select the distractors. Another manner is to apply some pre-defined rules to prepare distractors by changing the surface form of some words or phrases \cite{Chen2006FASTA}.
Automatic DG for RACE-like MCQs is a challenging task.
First, different from previous works that prepare word or short phrase distractors (1.46 tokens on average in SciQ \cite{Welbl2017CrowdsourcingMC}), we here endeavor to generate longer and semantic-rich distractors. Specifically, the average length of the distractors in our experimental dataset is 8.1.
Furthermore, the generated distractors should semantically related to the reading comprehension question, since it is trivial to identify a distractor having no connection with the article or question.
Moreover, the distractors should not be paraphrases of the correct answer option.
Finally, the generated distractors should be grammatically consistent with the question, especially for questions with a blank in the end, as shown in Figure~\ref{fig:example}. Previous works following the extraction-selection manner cannot meet these requirements.

We formulate the task of automatic distractor generation as a sequence-to-sequence learning problem that directly generates the distractors given the article, and a pair of question and its correct answer.
We design our framework to explicitly tackle the above mentioned challenges by using a data-driven approach to learn to meet these requirements automatically.
More specifically, we employ the hierarchical encoder-decoder network, which has already shown potentials to tackle long sequential input \cite{Tan2017FromNS,Ling2017CoarsetoFineAM}, as the base model for building our framework.
On top of the hierarchical encoding structure, we propose the dynamic attention mechanism to combine sentence-level and word-level attentions varying at each recurrent time step to generate a more readable sequence.
Furthermore, a static attention mechanism is designed to modulate the dynamic attention not to focus on question-irrelevant sentences or sentences which contribute to the correct answer option.
Finally, we use a question-based initializer as the start point to generate the distractor, which makes the distractor grammatically consistent with the question. In the generation stage, we use the beam search to generate three diverse distractors by controlling their distance.

In the evaluations, we conduct experiments on a distractor generation dataset prepared from RACE using n-gram based automatic evaluation metrics such as BLEU and ROUGE. The results show that our proposed model beats several baselines and ablations. Human evaluations show that distractors generated by our model are more likely to confuse the examinees, which demonstrates the functionality of our generated distractors in real examinations.
We will release the prepared dataset and the code of our model to facilitate other researchers to do further research along this line \footnote{Our code and data are available at \url{https://github.com/Evan-Gao/Distractor-Generation-RACE}}.

\section{Framework Description}

\subsection{Task Definition}
In the task of automatic Distractor Generation (DG), given an article, a pair of question and its correct option originated from the article, our goal is to generate context and question related, grammatically consistent wrong options, i.e. distractor, for the question.

Formally, let $P$ denote the input article containing multiple sentences: ${s_1, s_2, ..., s_n}$, $q$ and $a$ denote the question and its correct answer, respectively. The DG task is defined as finding the distractor $\overline{d}$, such that:
\begin{equation}\label{eqn.loglike}
\overline{d} = \argmax_{d} \text{log P(}d|P, a, q\text{)},
\end{equation}
where $\text{log P(}d|P, a, q\text{)}$ is the conditional log-likelihood of
the predicted distractor $d$, give $P$, $a$ and $q$.

\begin{figure*}[th]
    \centering
    \includegraphics[width=1.0\textwidth]{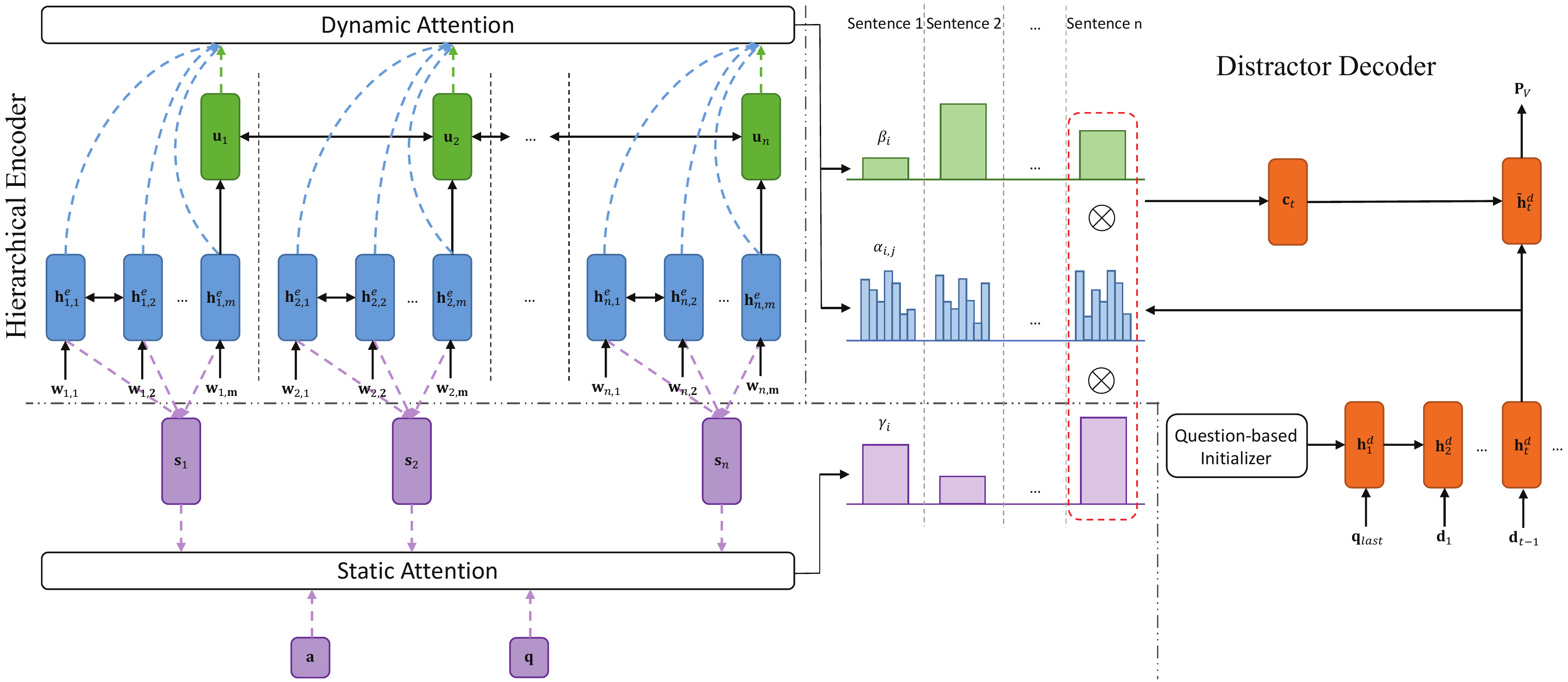}
    \caption{A overview of our model that jointly utilizes static and dynamic attentions.  \textit{(Better viewed in color).}}
    \label{fig:framework}
\end{figure*}

\subsection{Framework Overview}
A straightforward strategy for distractor generation is to employ the standard sequence-to-sequence learning network \cite{Sutskever2014SequenceTS} to learn the mapping from the article to the distractor. Unfortunately, an article can be too long as the input, which cannot receive decent results. Here we advocate the hierarchical encoder-decoder framework to model such long sequential input.
The architecture of our overall framework is depicted in Figure \ref{fig:framework}.

First, we employ the hierarchical encoder to obtain hierarchical contextualized representations for the whole article, namely, word-level representation and sentence-level representation. 
Before decoding the encoded information, we design a static attention mechanism to model the global sentence importance considering the fact that the distractor should be semantically related to the question and should not share the same semantic meaning with the correct answer. The static attention distribution is used in the decoder as a soft gate to modulate the dynamic attention.
For the decoder part, we first employ a language model to compress the question information into a fixed length vector to initialize the decoder state, making the distractor grammatically consistent with the question. During each decoding step, the dynamic hierarchical attention combines the sentence-level and word-level information to attend different part at each decoding time step.
With the combined architecture, our model can generate grammatically consistent, context and question related wrong options (distractors) in an end-to-end manner. 

\subsection{Hierarchical Encoder}
\subsubsection{Word Embedding.}
An embedding lookup table is firstly used to map tokens in each sentence $s_i$ in the article $P$ into word dense vectors $(\mathbf{w}_{i,1}, \mathbf{w}_{i,2}, ..., \mathbf{w}_{i,m})$, where $\mathbf{w}_{i,j}\in{\mathbb{R}^{d_w}}$ having $d_w$ dimensions. 

\subsubsection{Word Encoder.}
For each sentence $s_i$, the word encoder takes its word vectors $(\mathbf{w}_{i,1}, \mathbf{w}_{i,2}, ..., \mathbf{w}_{i,m})$ as input. Specifically, we use bidirectional LSTMs to encode the sequence to get a contextualized representation for each word:
\begin{equation}\label{eqn.word_enc}
\overrightarrow{\mathbf{h}^{e}_{i,j}} = \overrightarrow{\text{{LSTM}}}(\overrightarrow{\mathbf{h}^{e}_{i,j-1}}, \mathbf{w}_{i,j}),~ \overleftarrow{\mathbf{h}^{e}_{i,j}} = \overleftarrow{\text{{LSTM}}}(\overleftarrow{\mathbf{h}^{e}_{i,j+1}}, \mathbf{w}_{i,j}), \nonumber
\end{equation}
where $\overrightarrow{\mathbf{h}^{e}_{i,j}}$ and $\overleftarrow{\mathbf{h}^{e}_{i,j}}$ are the hidden states at the  $j$-th time step of the forward and the backward LSTMs. 
We concatenate them together as $\mathbf{h}^{e}_{i,j}=[\overrightarrow{\mathbf{h}^{e}_{i,j}};\overleftarrow{\mathbf{h}^{e}_{i,j}}]$. 

\subsubsection{Sentence Encoder.}
On top of the word encoding layer, we combine the final hidden state of the forward LSTM and the first hidden state of the backward LSTM of each sentence as the sentence representation and employ another bidirectional LSTMs to learn the contextual connection of sentences. We denote the contextualized representation of the sentence sequence as $(\mathbf{u}_1, \mathbf{u}_2, ..., \mathbf{u}_n)$.

\subsection{Static Attention Mechanism}
Recall that the generated distractors should be semantically relevant to the question, but must not share the same semantic meaning with the answer.
To achieve this goal, here we introduce a static attention mechanism which learns an importance distribution $(\gamma_1, \gamma_2, ..., \gamma_n)$ of the sentences $(s_1, s_2, ..., s_n)$ in the article.
Here we use the answer $a$ and the question $q$ as queries to interact with all sentences to learn such distribution.

\subsubsection{Encoding Layer.}
In the encoding layer, we transform the answer $a$, the question $q$ and all sentences $(s_1, s_2, ..., s_n)$ into fixed length vector representations. 
Specifically, two individual bidirectional LSTM networks are employed to encode $a$ and $q$ separately to derive the contextualized representation for each token in them and obtain $(\mathbf{a}_1, \mathbf{a}_2, ..., \mathbf{a}_k)$ and $(\mathbf{q}_1, \mathbf{q}_2, ..., \mathbf{q}_l)$, respectively.
Then an average pooling layer is employed to acquire the representation for the question and answer:
\begin{equation}\label{eqn.match_enc_wq}
\mathbf{a} = \frac{1}{k}\sum_{t=1}^{k} \mathbf{a}_t, \mathbf{q} = \frac{1}{l}\sum_{t=1}^{l} \mathbf{q}_t.
\end{equation}
For the sentence representation, we do not reuse the sentence representation $\mathbf{u}_i$ from the sentence encoder since $\mathbf{u}_i$ is responsible for learning the semantic information for a whole sentence, while here we only want to learn the importance distribution of sentences according to the query (i.e. a pair of question and answer).
Therefore, we only reuse the word-level contextualized representations $\mathbf{h}^{e}_{i,j}$ learned in the hierarchical encoder and employ the same average pooling layer to get the representation of each sentence:
\begin{equation}\label{eqn.match_enc_sent}
\mathbf{s}_i = \frac{1}{m}\sum_{t=1}^{m} \mathbf{h}^{e}_{i,t}.
\end{equation}

\subsubsection{Matching Layer.}
For generating non-trivial distractors, we should emphasize the sentences that are relevant to the question, and suppress the sentences relevant to the answer. For this reason, we learn a score $o_i$ for  $s_i$ that combines the above two aspects with bilinear transformation similar to \cite{Chen2018TitleGuidedEF}:
\begin{equation}\label{eqn.match_score}
o_i = \lambda_q \mathbf{s}_i^\top \mathbf{W}_m \mathbf{q} - \lambda_a \mathbf{s}_i^\top \mathbf{W}_m \mathbf{a} + \mathbf{b}_m,
\end{equation}
where $\mathbf{W}_m$ and $\mathbf{b}_m$ are learnable parameters.

\subsubsection{Normalization Layer.}
Before feeding the raw sentence importance score $o_i$ into the Softmax function to compute the final static attention distribution, we use the question to learn a temperature $\tau \in (0,1)$:
\begin{equation}\label{eqn.tau} \tau = \text{sigmoid}(\mathbf{w_q}^\top \mathbf{q} + b_q),
\end{equation}
where $\mathbf{w_q}$ and $b_q$ are learnable parameters. Then, we derive the static attention distribution as:
\begin{equation}\label{eqn.norm}
\gamma_i = \text{softmax}(o_i/\tau).
\end{equation}

The intuition behind using the temperature $\tau$ is that if a question asks for some specific details in the article, it is only relevant to one or two sentences. While if a question requires summarizing or reasoning, it could be relevant to many sentences in the article.
Therefore, we propose the above data-driven approach to learn the temperature $\tau$ according to the property of the question. If $\tau$ is close to 0, then it works together with $o_i$ to yield a peaked distribution $\gamma$ which simulates the case of detailed questions. Otherwise, if $\tau$ is close to 1, it will not peak any sentence attention score $\gamma_i$.

\subsection{Distractor Decoder}
We use another LSTMs as the decoder to generate the distractor. 
Instead of using the last hidden state of the encoder to initialize the decoder, we design a special question-based initializer to make the distractor grammatically consistent with the question.
During the decoding, we introduce the dynamic attention mechanisms to combine the sentence-level and word-level attentions varying at each recurrent time step to generate a more readable sequence. We also incorporate the static attention here to modulate the dynamic attention to ensure the semantic relevance of the generated distractors.

\subsubsection{Question-based Initializer.}
We design a question-based initializer to initialize the initial state of the decoder.
Specifically, we use a question LSTM to encode the question, and use the last time step information of the LSTM in the following manner:
\begin{itemize}
    \item Instead of using BOS (i.e. the \textit{Begin of Sentence} indicator), we use the last token in the question ($\mathbf q_{last}$) as the initial input of the decoder.
    \item Other than using the final state of the hierarchical encoder to initialize the decoder, we here use the final cell state and hidden state of the question LSTM to initialize the decoder.
\end{itemize}

\subsubsection{Dynamic Hierarchical Attention Mechanism.}
The standard attention mechanism treats an article as a long sequence and compares the hidden state of the current decoding time step to all encoder hidden states.
This approach is not suitable for long input sequences for the following reasons. First, the standard LSTM cannot model such long inputs (on average, 343.9 words per article in our training set). Second, we will lose the sentence structure if we treat the tokens of different sentences equally. Last but not least, usually a question or a distractor is only related to a small number of sentences in the article, we should only use the related sentences to generate the distractor, but the standard attention has no emphasis on difference sentences.

Given the above reasons, we employ the dynamic hierarchical attention to only focus on important sentences during each decoding time step. We call it \textit{dynamic} because both word-level and sentence-level attention distributions change at each time step.
When generating a word at the time step $t$, the decoder reads the word embedding $\mathbf{d}_{t-1}$ and the hidden state $\mathbf{h}^{d}_{t-1}$ of the previous time step to generate the current hidden state $\mathbf {h}^{d}_t = \text{LSTM}(\mathbf {h}^{d}_{t-1}, \mathbf d_{t-1})$.
Then it calculates both the sentence-level attention $\beta_i$ and the word-level attention $\alpha_{i,j}$ at the same time:
\begin{equation}\label{eqn.dyn_att}
\beta_i = \mathbf{u}_i^\top \mathbf{W}_{d_1} \mathbf{h}^{d}_t, ~~~ \alpha_{i,j} = {\mathbf{h}^{e}_{i,j}}^\top \mathbf{W}_{d_2} \mathbf{h}^{d}_t,
\end{equation}
where $\mathbf{W}_{d_1}$ and $\mathbf{W}_{d_2}$ are trainable parameters.
The sentence-level attention determines how much each sentence should contribute to the generation at the current time step, while the word-level attention determines how to distribute the attention over words in each sentence.

Finally, we use the static attention $\gamma_i$ to modulate the dynamic hierarchical attention $\beta_i$ and $\alpha_{i,j}$ by simple scalar multiplication and renormalization. Thus, the combined attention for each token in the article is:
\begin{equation}\label{eqn.comb_att}
 \widetilde{\alpha}_{i,j} = \frac {\alpha_{i,j} \beta_i \gamma_i} {\sum_{i，j} \alpha_{i,j} \beta_i \gamma_i}.
\end{equation}
Then the context vector $\mathbf c_t$ is derived as a combination of all article token representations reweighted by the final combined attention $\widetilde{\alpha}_{i,j}$:
\begin{equation}\label{eqn.dec_context}
\mathbf{c}_t = \sum_{i,j}\widetilde{\alpha}_{i,j} \mathbf{h}^{e}_{i,j}.
\end{equation}
And the attentional vector is calculated as: 
\begin{equation}
\tilde{\mathbf{h}}^{d}_t = \text{tanh}(\mathbf{W}_{\tilde{\mathbf{h}}} [\mathbf{h}^{d}_t;\mathbf{c}_t]).
\end{equation}
Then, the predicted probability distribution over the vocabulary $V$ at the current step is computed as:
\begin{align}\label{eqn.dec_out}
\text{P}_{V} = \text{softmax}(\mathbf{W}_V \tilde{\mathbf{h}}^{d}_t + \mathbf{b}_V),
\end{align}
where $\mathbf{W}_{\tilde{\mathbf{h}}}$, $\mathbf{W}_V$ and $\mathbf{b}_V$ are learnable parameters.

\subsection{Training and Inference}
Given the training corpus $\mathcal{Q}$ in which each data sample contains a distractor $d$, an article $P$, a question $q$ and an answer $a$, we minimize the negative log-likelihood with respect to all learnable parameters $\Theta$ for training:
\begin{align}\label{eqn.train}
\mathcal{L} =  - \sum_{d \in \mathcal{Q}} \text{log P(}d|P,a,q; \Theta \text{)}.
\end{align}
During generation, if UNK (i.e. unknown words) is decoded at any time step, we replace it with the word having the largest attention weight in the article.

Since there are several diverse distractors (2.4 on average according to Table \ref{tab:stat}) corresponding to the same question in our dataset, we use beam search with beam size $k$ in the testing stage and receive $k$ candidate distractors with decreasing likelihood. 
The ultimate goal is to generate several diverse distractors, however, usually the successive output sequences from beam search would be similar. Therefore we design the following protocol to generate three diverse distractors. Firstly, we select the distractor with the maximum likelihood as $d^{g}_1$. Then we select $d^{g}_2$ among the remaining candidate distractors along the decreasing order of the likelihood, restricting that the Jaccard distance between $d^{g}_1$ and $d^{g}_2$ is larger than 0.5. Finally, $d^{g}_3$ is selected in a similar way where its distances to both of $d^{g}_1$ and $d^{g}_2$ are restricted.

\section{Experimental Settings}


\begin{table}[!t]
    \centering
    {\small
    \begin{tabular}{lc} 
    \Xhline{2\arrayrulewidth}
    \hline
    {\# Train Samples}   & 96501    \\ 
    {\# Dev Samples}   & 12089     \\ 
    {\# Test Samples}   & 12284     \\ 
    \hline
    {Avg. article length (tokens)} & 347.0  \\
    {Avg. distractor length} & 8.5  \\
    {Avg. question length} & 9.9  \\
    {Avg. answer length} & 8.7  \\
    \hline
    {Avg. \# distractors per question} & 2.1 \\
    \hline
    \Xhline{2\arrayrulewidth}
    \end{tabular}
    }
    \caption{The statistics of our dataset.}
    \label{tab:stat}
\end{table}

\subsection{Dataset}
We evaluate our framework on a distractor generation dataset prepared with the RACE \cite{Lai2017RACELR} dataset.
RACE contains 27,933 articles with 97,687 questions from English examinations of Chinese students from grade 7 to 12.
We first extract each data sample as a  quadruple of article, question, answer and distractor from RACE, followed by some simple preprocessing steps, such as tokenization, sentence splitting, and lower-casing.

After some investigation on the RACE dataset, we observe that some distractors have no semantic relevance with the article, which can be easily excluded in the examination and also do not make sense for the task of distractor generation since our goal is to generate confusing distractors. Hence, we first filter out such irrelevant distractors by simply counting meaningful tokens in individual distractors. We define a token meaningful if it is not a stop word and has a POS tag from \{`JJ', `JJR', `JJS', `NN', `NNP', `NNPS', `NNS', `RB', `RBR', `RBS', `VB', `VBD', `VBG', `VBN', `VBP', `VBZ'\}. Then, we prune the dataset based on the following constraint: For those meaningful tokens in a distractor that also appear in the article, if their total weighted frequency is no less than 5, the distractor will be kept. Here the weighted frequency of a meaningful token means the multiplication of its frequency in the distractor and its frequency in the article. 
Moreover, we remove the questions which need to fill in the options at the beginning or in the middle of the questions. Table \ref{tab:stat} reports the statistics of the processed dataset. We randomly divide the dataset into the training (80\%), validation (10\%) and testing sets (10\%). 

\subsection{Implementation Details}
We keep the most frequent 50k tokens in the entire training corpus as the vocabulary, and use the \texttt{GloVe.840B.300d} word embeddings \cite{Pennington2014GloveGV} for initialization and finetune them in the training. 
Both source and target sides of our model share the same word embedding. All other tokens outside the vocabulary or cannot found in GloVe are replaced by the UNK symbol.
We set the number of layers of LSTMs to 1 for the hierarchical encoder (for both word encoder and sentence encoder) and the static attention encoder, and 2 for the decoder. The bidirectional LSTMs hidden unit size is set to 500 (250 for each direction).
For the LSTM used in the question-based  initialier, we use 2 layers unidirectional LSTMs with hidden size 500.
The hyperparameters $\lambda_q$ and $\lambda_a$ in static attention are initialized as 1.0 and 1.5 respectively.
We use dropout with probability $p=0.3$. 
All trainable parameters, except word embeddings, are randomly initialized with $\mathcal{U}(-0.1, 0.1)$. 
For optimization in the training, we use stochastic gradient descent (SGD) as the optimizer with a minibatch size of 32 and the initial learning rate 1.0 for all baselines and our model. We train the model for 100k steps and start halving the learning rate at step 50k, then we halve the learning rate every 10k steps till ending. We set the gradient norm upper bound to 5 during the training.
We employ the teacher-forcing training, and in the generating stage, we set the maximum length for output sequence as 15 and block unigram repeated token, the beam size $k$ is set to 50. 
All hyperparameters and models are selected on the validation set based on the lowest perplexity and the results are reported on the test set.

\begin{table*}[!t]
\centering
\begin{tabular}{ccccccccc}
\Xhline{3\arrayrulewidth}
                     &      & BLEU$_{1}$    & BLEU$_{2}$    & BLEU$_{3}$   & BLEU$_{4}$   & ROUGE$_{1}$ & ROUGE$_{2}$ & ROUGE$_{L}$ \\ \hline
\multirow{3}{*}{1st Distractor}   & Seq2Seq  & 25.28 & 12.43 & 7.12 & 4.51 & 14.12 & 3.35  & 13.58 \\
                     & HRED & 26.10 & 13.96 & 8.83 & 6.21 & 14.83 & 4.07  & 14.30 \\ 
                     & Our Model  & \textbf{27.32} & \textbf{14.69} & \textbf{9.29} & \textbf{6.47} & \textbf{15.69} & \textbf{4.42}  & \textbf{15.12} \\ \hline
\multirow{3}{*}{2nd Distractor}   & Seq2Seq  & 25.13 & 12.02 & 6.56 & 3.93 & 13.72 & 3.09  & 13.20 \\
                     & HRED & 25.18 & 12.21 & 6.94 & 4.40 & 13.94 & 3.11  & 13.40 \\ 
                     & Our Model  & \textbf{26.56} & \textbf{13.14} & \textbf{7.58} & \textbf{4.85} & \textbf{14.72} & \textbf{3.52}  & \textbf{14.15} \\ \hline
\multirow{3}{*}{3rd Distractor}   & Seq2Seq  & 25.34 & 11.53 & 5.94 & 3.33 & 13.78 & 2.82  & 13.23 \\
                     & HRED & 25.06 & 11.69 & 6.26 & 3.71 & 13.65 & 2.84  & 13.04 \\ 
                     & Our Model  & \textbf{26.92} & \textbf{12.88} & \textbf{7.12} & \textbf{4.32} & \textbf{14.97} & \textbf{3.41}  & \textbf{14.36} \\ \hline
\multirow{3}{*}{Avg. Performance} & Seq2Seq  & 25.25 & 11.99 & 6.54 & 3.92 & 13.87 & 3.09  & 13.34 \\
                     & HRED & 25.45 & 12.62 & 7.34 & 4.77 & 14.14 & 3.34  & 13.58 \\ 
                     & Our Model  & \textbf{26.93} & \textbf{13.57} & \textbf{8.00} & \textbf{5.21} & \textbf{15.13} & \textbf{3.78}  & \textbf{14.54} \\ 
\Xhline{3\arrayrulewidth}
\end{tabular}
\caption{Automatic evaluation results on all systems by BLEU and ROUGE.
1st, 2nd and 3rd distractors are generated under the same policy.
The best performing system for each compound row is highlighted in boldface.}
\label{tab:auto}
\end{table*}

\subsection{Baselines and Ablations}
We compare our framework with the following baselines and ablations. \textbf{Seq2Seq}: the basic encoder-decoder learning framework \cite{Sutskever2014SequenceTS} with attention mechanism \cite{Luong2015EffectiveAT}. Here we adopt the global attention with general score function. The hidden size of LSTMs for both encoder and decoder is 500. We select the model with the lowest perplexity on the validation set. \textbf{HRED}: the \textbf{H}ie\textbf{R}archical \textbf{E}ncoder-\textbf{D}ecoder~(HRED) with hierarchical attention mechanism. This architecture has been proven effective in several NLP tasks including summarization \cite{Ling2017CoarsetoFineAM}, headline generation \cite{Tan2017FromNS}, and text generation \cite{Li2015AHN}. Here we keep the LSTMs size as 500 for fairness and set the number of the word encoder and sentence encoder layers as 1 and the decoder layer as 2. 
We employ the question-based initializer for all baselines to generate grammatically coherent distractors. In the generation stage, we follow the same policy and beam size for baselines and ablations during the inference stage to generate three distractors.

\section{Results and Analysis}

\subsection{Automatic Evaluation}
Here we evaluate the similarity of generated distractors with the ground truth. We employ BLEU (1-4) \cite{Papineni2002BleuAM} and ROUGE (R1, R2, R-L) \cite{Lin2004ROUGEAP} scores to evaluate the similarity. BLEU evaluates average n-gram precision on a set of reference sentences, with a penalty for overly long sentences. $\text{ROUGE}_{1}$ and $\text{ROUGE}_{2}$ is the recall of unigrams and bigrams while $\text{ROUGE}_{L}$ is the recall of longest common subsequences.

Table \ref{tab:auto} shows the automatic evaluation results of all systems. Our model with static and dynamic attentions achieve the best performance across all metrics. 
We can observe a large performance gap between Seq2Seq and models with hierarchical architectures (HRED and our model), which reveals the hierarchical structure is useful for modeling the long sequential input. Another reason could be that some distractors can be generated only use information in several sentences, and sentence-level attentions (both static and dynamic) are useful to emphasize several sentences in the article. Moreover, our model with static attention achieves better performance than its ablation HRED, which shows the static attention can play the role of a soft gate to mask some irrelevant sentences and modulate the dynamic attention.

By comparing the three distractors generated by beam search with a predefined Jaccard distance, we find that the performance drops a little for the second and third distractors. The reason can be two-folds: 1) The second and third distractors have lower likelihood; 2) We set a Jaccard distance threshold as 0.5 to select the second and third distractors, thus they are forced to use some words different from those in the first distractor which is likely to be the best generation. 

It is worth to mention that another automatic evaluation method can be applying a state-of-the-art reading comprehension pipeline for RACE to test its performance on our generated distractors. However, the current best performance of such reading comprehension pipeline is only 53.3\% \cite{Wang2018ACM,zhu2018hierarchical,Xu2017DynamicFN,Tay2018MultirangeRF}, which means half questions in the dataset cannot be answered correctly. Therefore, we do not employ such reading comprehension pipeline to evaluate our generated distractors, instead we hire human annotators to conduct a reliable evaluation, given in the next section.

\begin{figure*}[th]
    \centering
    \includegraphics[width=0.9\textwidth]{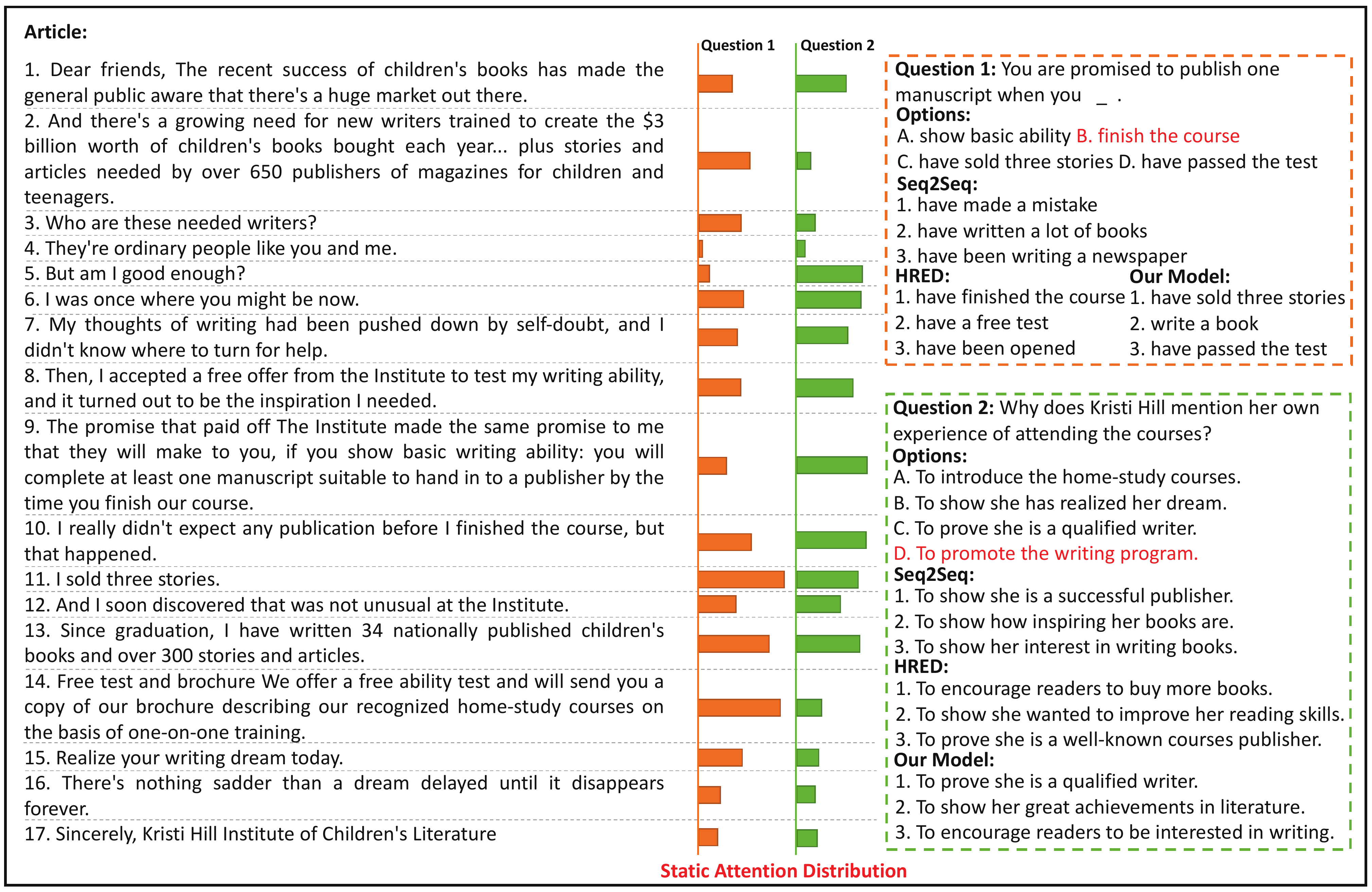}
    \caption{Sample generated distractors. On the right, two example questions are given in dotted lines of yellow and green, and their corresponding static attention distributions are given in the middle by the bars of the corresponding colors. }
    \label{fig:case}
\end{figure*}

\subsection{Human Evaluation}
\begin{table}
    \centering
    {
    \resizebox{1.0\columnwidth}{!}{
    \begin{tabular}{c|cccc} 
    \Xhline{2\arrayrulewidth}
    \hline
     & Annotator 1 & Annotator 2 & Annotator 3 & \# Selected    \\ 
    \hline
    {Seq2Seq} & 31 & 35 & 30 & 96     \\ 
    {HRED} & 33 & 40 & 35 & 108     \\ 
    {Our Model} & 43 & 45 & 36 & 124  \\
    \hline
    {Human} & 75 & 70 & 79 & 224  \\
    \hline
    \Xhline{2\arrayrulewidth}
    \end{tabular}}
    }
    \caption{Human evaluation results. Note that we allow annotators to choose more than one options if the generated outputs are accidentally the same or very semantically similar, therefore, the total number of selected options (552) is larger than the total number of annotated questions (540).}
    \label{tab:manual}
\end{table}

We conduct a human evaluation to investigate if the generated distractors can confuse the examinees in the real human test. We employ three annotators with good English background (at least holding a bachelor degree) to answer the MCQs with the generated distractors from different methods. Specifically, for each MCQ, we give 4 distractors as its options: One is a sample from the ground truth, the other three are generated by Seq2Seq, HRED, and our model respectively. Note that we did not give the correct answer option to the annotators, because the current human ceiling performance on RACE dataset is about 95\%~\cite{Lai2017RACELR}. Thus, we need to do a huge amount of annotation for collecting enough questions that are answered wrongly. During the annotation, we told the annotators to select the most suitable option without considering whether there exists a correct option.

For comparison, we count how many times of individual pipelines (the ground truth and three compared methods) are successful in confusing the annotators, i.e. their distractors are selected as answers. We give each annotator 60 articles, and 3 questions per article. In total, we annotated 540 questions, and the results are given in Table \ref{tab:manual}. We find that the ground truth distractors (i.e. by ``Human'') have the strongest capability to confuse the annotators. Among the compared automatic methods, our model performs the best, while Seq2Seq performs the worst, which is a consistent conclusion as drawn from the previous section. 


\subsection{Case Study}

In Figure \ref{fig:case}, we present some sample distractors generated by human instructors, the Seq2Seq baseline, HRED and our model. To validate the effectiveness of the static attention, we show the static attention distributions over the sentences of the article for the two example questions. 
The correct options of the questions are marked in red.

Question 1 asks a detailed aspect in the article, which can be directly answered according to the 9th sentence. Since our static attention mechanism suppresses the sentences which contain the answer information, we can see the score for the 9th sentence is relatively smaller than others.
The distractor outputs also justify our intuitions. Specifically, the first distractor by \textbf{HRED} is semantically identical to the correct option, thus it is not an appropriate distractor. With the help of the static attention, our model does not generate distractors like this.
Another effect of the static attention is that it highlights the sentences that are relevant to the question, such as 11th, 13th, and 14th sentences, so that our model can generate better distractors. We can see the distractors generated by our model are semantically relevant to these highlighted sentences. 
Last but not least, we find that the distractors generated by Seq2Seq baseline either focus on some frequent words in the article such as \textit{publish} and \textit{write}, or contain some completely irrelevant words such as \textit{mistake} and \textit{newspaper}.
HRED and our model do not have this problem, because the dynamic hierarchical attention can modulate the word-level attention distribution with the sentence-level attention.

By looking at Question 2, we can also find that the distractors generated by our system are more appropriate and relevant. Because Question 2 requires some inference, it is thus relevant to several sentences across the article. The static attention distribution yields the same conclusion. Specifically, the distribution shows that the 5th to 13th sentences are all relevant to the question, while the 14th sentence which is relevant to the answer option is suppressed. The generated distractors from our system are also semantically relevant to the 5th to 13th sentences.


\section{Conclusions}
In this paper, we present a data-driven approach to generate distractors from multiple choice questions in reading comprehension from real examinations. We propose a hierarchical encoder-decoder framework with dynamic and static attention mechanisms to generate the context relevant distractors satisfying several constraints. We also prepare the first dataset for this new setting, and our model achieves the best performance in both automatic evaluation and human evaluation.
For the future work, one interesting direction is to transform this one-to-many mapping problem into one-one mapping problem to better leverage the capability of the sequence-to-sequence framework. 
Another promising direction could be explicitly adding supervision signals to train the static attention.
From the perspective of RACE-like reading comprehension tasks with multiple choice questions, although the performance of existing reading comprehension methods are still quite unsatisfactory, by introducing the distractor generation task, it might open another door for improving the performance, i.e. making adversarial approaches for solving this reading comprehension task possible.

\section{Acknowledgments}
The work described in this paper was supported by the Research Grants Council of the Hong Kong Special Administrative Region, China (No. CUHK 14208815 and No. CUHK 14210717 of the General Research Fund), and Microsoft Research Asia (2018 Microsoft Research Asia Collaborative Research Award).

\bibliography{aaai19}
\bibliographystyle{aaai}

\end{document}